# Multi-view knowledge distillation transformer for human action recognition

Ying-Chen Lin[a], Vincent S. Tseng[b]

[a] Institute of Computer Science and Engineering, National Yang Ming Chiao Tung University, Hsinchu 30010, Taiwan.

[b] Department of Computer Science, National Yang Ming Chiao Tung University, Hsinchu 30010, Taiwan.

## Abstract

Recently, Transformer-based methods have been utilized to improve the performance of human action recognition. However, most of these studies assume that multi-view data is complete, which may not always be the case in real-world scenarios. Therefore, this paper presents a novel Multi-view Knowledge Distillation Transformer (MKDT) framework that consists of a teacher network and a student network. This framework aims to handle incomplete human action problems in real-world applications. Specifically, the multi-view knowledge distillation transformer uses a hierarchical vision transformer with shifted windows to capture more spatial-temporal information. Experimental results demonstrate that our framework outperforms the CNN-based method on three public datasets.

## 1. Introduction

Human action recognition is popular in computer vision. However, single-view human action recognition is difficult to be applied in real life because we cannot control the free movement of people, so we can capture more information through shooting from multiple perspectives, in addition when some shooting equipment is damaged, it can also maintain a certain performance of recognition.

The research field of Multi-view Human Action Recognition can be divided into two categories. The first category is Multi-viewpoints, which are taken from different angles simultaneously. The second category is Multi-modality, which uses the same shooting angle but contains more than two modalities.

In the previous research, multi-viewpoint characteristics were learned through different modalities, such as optical flow[5]. For example, [1] uses the pre-trained model to generate DensePose modality in advance and then uses DensePose and RGB modality during training, and [2] uses RGB modality to generate Depth modality and then uses these two modalities to train the model. The common shortcoming of these studies is that they all need to spend extra time to collect another modality. Although they use different modalities to improve performance, they ignore that these modalities cannot be easily obtained in real life, so we decided only to use the RGB modality to simulate the incomplete data situation.

In addition, some work[3, 10-16] used transformer modules to learn human action recognition in recent years. Therefore, we design a transformer-based module for our framework.

To recap, the main contributions of our proposed method are summarized as follows.

We propose a prediction framework called Multi-View Knowledge Distillation Transformer (MKDT), which requires only RGB modality for human action recognition during the learning phase, thereby reducing equipment costs. Our framework is based on a transformer, which predicts human action recognition on incomplete viewpoints and outperforms CNN-based (Convolutional Neural Network) models. We expect our framework to achieve state-of-the-art performance with only a single view of inference.

## 2. Related work

### 2.1 Human Action Recognition

Human action recognition has recently become a popular research topic in computer vision [6, 7]. As self-attention[8] in natural language processing has attracted more and more attention, self-attention technology has also begun to be introduced in computer vision. Many previous works [9-16] have used self-attention.

### 2.2 Multi-view Action Recognition

Multi-viewpoint action recognition aims to shoot through multiple cameras simultaneously, which can effectively integrate much information from different aspects[4, 17-26]. First, [17, 18] Zheng et al. present a method for view-invariant action recognition based on sparse representations using a transferable dictionary pair. Even though the above approaches have achieved promising results, they ignore the interactive relatedness among multiple actions from multiple views. To handle this problem, Hao et al. [19] proposed bad-of-visual-words (BoVW)/fisher vector(FV) to represent individual videos in each view. Then, the sparse coding algorithm transfers features of various views into the discriminative and high-level semantics space. Finally, they employ the multi-task learning (MTL) approach for joint action modeling and discovering latent relationships among different action categories.

Recently year, Wang et al. [20] designed a Dividing and Aggregating Network (DA-Net) to learn view-independent representations shared by all views at lower layers. In comparison, they learn one view-specific representation for each view at higher layers. Then, they train view-specific action classifiers based on the view-specific representation for each view and a view classifier based on the shared representation at lower layers. Vyas et al. [21] propose a framework for unsupervised multi-view representation learning via cross-view video prediction, which is integrated to get a holistic representation of the action from multiple views. This allows the network to preserve the notion of viewpoint and time, which facilitates query-based cross-view prediction. Jiang et al. [22] proposed a deep probabilistic model to quantify multi-view social behavior in mice. Their approach jointly models the temporal relationship of frames in each view, the relationship between views, and the correlation between labels in the neighboring areas. Ullah et al. [23] proposed a conflux structure of the LSTMs network, which has separate LSTM for each view and processes the sequential features obtained from the consecutive frames. This structure allows their network to learn the view self-reliant sequential patterns by processing single-view data. Schmidt et al. [24] implement an attention-based model to mix multiple views into a global representation. Liu et al. [25] use multi-view information to reconstruct the daily action recognition scheme. For the unsupervised segmentation of multi-view motion atoms, they introduce a task-based dictionary to obtain a more discriminative classifier, which can effectively improve the classification accuracy of motion atoms. Shan et al. [26] proposed a novel multi-view driver action recognition architecture named Multi-view Vision transformer(MVVT), which combines a standard 2D CNN encoder with a modified vision transformer.

## 2.3 Multi-modality Action Recognition

Multi-modality action recognition aims at recognizing human activity using multiple modalities, such as depth, RGB, infrared, skeleton, acceleration, etc., which can effectively integrate much information from different aspects[1、2、27-29].

Recently works used multi-modalities for action recognition to improve performance. Wang et al. [27] proposed a Generative Multi-View Action Recognition (GMVAR) framework to address incomplete multi-view data problems. The adversarial generative network is leveraged to generate one view conditioning on the other view, which fully explores the latent connections in both intra-view and cross-view aspects. Lately, Liang et al. [2] observed that work late fusion could learn the final result without considering the correlation among the different models for each view. Attention has been widely adopted as an effective strategy for discovering discriminative cues underlying temporal data. Bai et al. [28].  proposed a collaborative attention mechanism (CAM) to detect the attention differences among multi-view. Specifically, they extend the long short-term memory (LSTM) to a Mutual-Aid RNN (MAR) to achieve the multi-view collaboration process.   Xiong et al. [29] present a multi-view pseudo-labeling approach to video learning. This novel framework uses complementary views in the form of appearance and motion information for semi-supervised learning in the video.

## 3   Proposed Framework

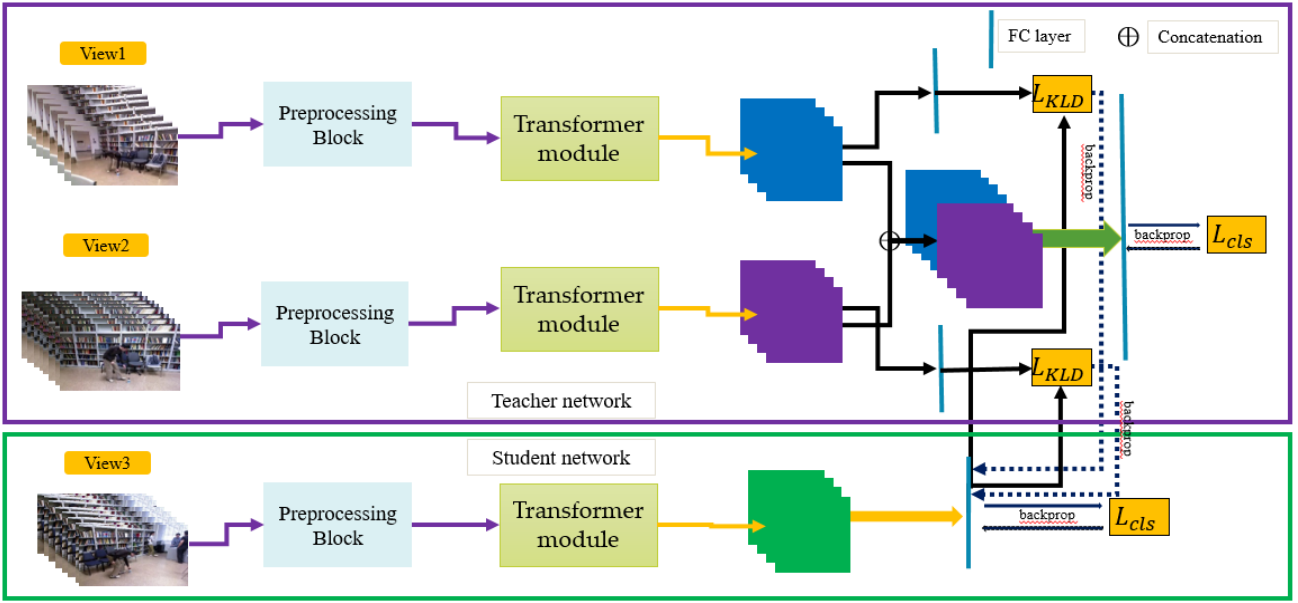

**Fig. 1.** The overview architecture of our framework. For example, three views and RGB information are used. The teacher network of our MKDT is given multi-view RGB sequences. Our method is designed in an online distillation manner.

### 3.1 Multi-View Knowledge Distillation Transformer

For the multi-view action recognition dataset $D$, it can divide $D_{train}$ and $D_{val}$ according to different and independent views. The training set $D_{train}$ contains video sequences of the same actor's action in several different views $\{V_1, V_2, …, V_k\}$, where k in a total number of views, and the validation set $D_{val}$ includes video sequences of a new view $V_i$.

The overall architecture of our framework is illustrated in **Fig. 1**. Let us denote the available training set as $\{(X_{v_1}, X_{v_2}, …, X_{v_k}); y\}$, where $X$ is the training sample from the $V_i$ view, y is score vector of classification. After learning, the network needs to map video sequences $\{(X_{v_1}, X_{v_2}, …, X_{v_k})\}$ into y.

In this work, we propose a teacher-student learning framework named a multi-view knowledge distillation transformer(MKDT). First, the original RGB sequences$\{(X_{v_1}, X_{v_2}, …, X_{v_k})\}$ addition to pre-processing block as a feature for model learning. Next, the normalized feature is fed into the transformer module to learn the representations. . Finally, we extract the feature maps from a fully connected layer.

### 3.2 Transformer module

In recent years, the vision community witnessing a modeling shift from CNNs to Transformers. Bertasius et al. [9] propose the first work that used self-attention in video action recognition. However, the effects of this work are poor to the used spatial-temporal information. Therefore, we used Video Swin Transformer[11] as our transformer module.

### 3.3 Multi-view knowledge transfer

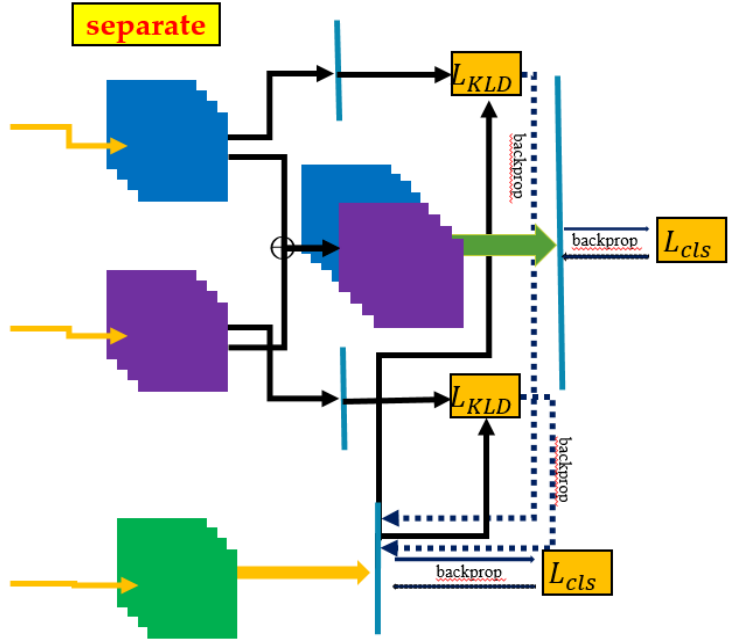

**Fig. 2.** Diagram of late fusion loss. As illustrated, the cross entropy loss calculates from the ground truth and models' clarification probability. The KL Divergence loss would separate calculated by feature maps between the teacher network and student network.

To leverage view knowledge from multiple views, in this section, we adopt a method of separate late fusion to calculate the distance of the representation between the teacher network and the student network. For multi-view learning, the initial result is learned by the forward inference on each view, and then the late fusion is used for the final result. In previous work[1, 27], direct fusion feature maps into the fully connected layers , which may impact the learning effectiveness of the model. Therefore, we propose a separate process to calculate the loss. As a result, we redefine the loss formulated, shown in **Fig. 2**. the total losses for training $L_{teacher}$ and $L_{student}^{separate}$ are formulated as follows.

$$L_{teacher} = L_{cls}^{teacher}$$

$$L_{student}^{separate} = L_{cls}^{student} + \gamma L_{KLD_{v1}}^{student} + \ldots + \gamma L_{KLD_{vn}}^{student}, \text{ where } \gamma=1, n \text{ is number of viewpoints} \quad (1)$$

## 4 Experiments

### 4.1 Datasets

IXMAS[30] is a multi-view human action recognition dataset. It contains 11 actions: punch, pick-up, point, waving, checking a watch, etc. The dataset is relatively small and has a total of 1800 labeled video clips. Each action is performed by ten actors three times, captured synchronously by four horizontal cameras and over vertical cameras.

N-UCLA[31] is another multi-view action recognition benchmark dataset. It contains ten daily actions captured by three static cameras and performed by ten subjects several times. The dataset consists of 1449 RGB videos, correlated depth frames, and skeleton information. We only use RGB videos in our experiments.

3MDAD[32] is the other multi-view action recognition dataset. It is a real-world public dataset that contains RGB frames and depth maps recorded from two views. More than 60 drivers are involved, and each of the subjects performs 16 different actions under the naturalistic driving condition. In total, the dataset consists of 1104 videos.

## 4.2 Experimental Settings

We used stratified k-fold cross-validation to evaluate our method performance, where k is set up five. To make our evaluation more in line with practical application scenarios, we used actors as the basis for splitting the training and validation sets.

We implement our multi-view action recognition transformer architecture in Pytorch. Our model is evaluated based on three benchmark datasets, IXMAS[30], N-UCLA[31], and 3MDAD[32]. We train and evaluated on a server with three NVIDIA GeForce GTX 2080 Ti GPUs.

Our framework is optimized by an AdamW[33] optimizer, and the learning rate is $1 \times 10^{-4}$. For each experiment, the network was trained with fifteen epochs. The batch size is set to 8.

## 4.3 Comparison with state-of-the-art

We compared the recognition accuracy of MKDT with a state-of-the-art model[1]. **Table 1** shows the accuracy of the N-UCLA dataset. Our method performs better than MFS on each view. In detail, when $Cam_1$, $Cam_2$, and $Cam_3$ are used as the test set separately, our method achieves an accuracy of 83.4, 69.4, and 73.0, which is 21.3%, 9.3%, and 14% higher than the MFS[1].

The experimental result of IXMAS dataset is reported in **Table 2**. In contrast, using the top view as testing could be given a higher performance improvement ratio than using the horizontal view when applying our method.

For the 3MDAD dataset experiment, we used a feature map of late joint fusion to calculate KL Divergence loss. The 3MDAD night dataset's performance improvement ratio is higher compared with the 3MDAD day datasets by using our method. In detail, when $Cam_1$ and $Cam_2$ are used as the test set separately, our method achieves an accuracy of 76.71, 75.71, which is 17.59% and 20.61% higher than the MFS[1]. The experimental result is shown in **Table 3**.

**Table 1**

Comparisons of testing accuracies of the MFS action classification on N-UCLA. $Cam_1$ indicates that $Cam_1$ is used as the test, while the other camera views are used as the train, and so on.

| Method | $Cam_1$ | $Cam_2$ | $Cam_3$ |
|---|---|---|---|
| **MFS[1]** | 62.10 | 60.10 | 59.00 |
| **MKDT(Ours)** | 83.38 | 69.34 | 73.00 |

**Table 2**

Comparisons of testing accuracies of the MFS action classification on IXMAS. $Cam_1$ indicates that $Cam_1$ is used as the test, while the other camera views are used as the train, and so on. $Cam_1$- $Cam_4$ are the horizontal views, and $Cam_5$ is the top view.

| Method | $Cam_1$ | $Cam_2$ | $Cam_3$ | $Cam_4$ | $Cam_5$ |
|---|---|---|---|---|---|
| **MFS[1]** | 80.00 | 83.89 | 79.44 | 81.94 | 59.44 |
| **MKDT(Ours)** | 86.95 | 86.67 | 88.05 | 86.11 | 71.95 |

**Table 3**

Comparisons of testing accuracies of the MFS action classification on 3MDAD. $Cam_1$ indicates that $Cam_1$ is used as the test, while the other camera views are used as the train, and so on.

| | Day | | Night | |
|---|---|---|---|---|
| Method | $Cam_1$ | $Cam_2$ | $Cam_1$ | $Cam_2$ |
| **MFS[1]** | 77.50 | 67.38 | 59.12 | 55.10 |
| **MKDT(Ours)** | 83.00 | 79.50 | 76.71 | 75.71 |

## 4.4 Compare with single-view

To explore the effect of multi-view, we experiment with single-view on three datasets, As **Table 4** shows, with individual-view input, obvious lower accuracies are obtained.

**Table 4**

Comparisons of testing accuracies with single-view and multi-view.

| Dataset | View | Single-view | Multi-view |
|---|---|---|---|
| **N-UCLA** | $Cam_1$ | 79.80 | 83.38 |
| | $Cam_2$ | 68.51 | 69.34 |
| | $Cam_3$ | 70.07 | 73.00 |
| **IXMAS** | $Cam_1$ | 82.78 | 86.95 |
| | $Cam_2$ | 82.50 | 86.67 |
| | $Cam_3$ | 83.89 | 88.05 |
| | $Cam_4$ | 82.76 | 86.11 |
| | $Cam_5$ | 68.33 | 71.95 |
| **3MDAD(Day)** | $Cam_1$ | 79.75 | 83.00 |
|  | $Cam_2$ | 74.25 | 79.50 |
| **3MDAD(Night)** | $Cam_1$ | 68.50 | 76.71 |
|  | $Cam_2$ | 66.35 | 75.71 |

## 5 Conclusions

In order to introduce the transformer in video human action recognition, this paper presents a multi-view knowledge distillation transformer that aims to learn more representations from multi-view. Specifically, we use a video swin transformer as the backbone of our framework to capture more spatial-temporal information. Experimental results demonstrate that our framework can improve the performance of multi-view human action recognition on three public datasets.